\documentclass[conference,10pt]{IEEEtran}
\usepackage{amsmath,amssymb,amsthm,mathrsfs,amsfonts,dsfont,stmaryrd, wasysym,marvosym}
\usepackage{mathtools}
\usepackage{color}
\usepackage{graphicx,subcaption,cite,epsfig,dblfloatfix,url,xcolor} 
\usepackage{etoolbox}
\allowdisplaybreaks[4] 
\usepackage{caption}

\newtheorem{remark}{\textbf{Remark}}
\newcommand{\dv}{\mathbf} % determenistic vector
\newcommand{\bm}{\mathbf} % determenistic vector
\newcommand{\mc}{\mathcal} % determenistic vector
\newcommand{\mkv}{-\!\!\!\!\minuso\!\!\!\!-}

\usepackage{makecell}
\usepackage{pxfonts}
\usepackage{dsfont}
\graphicspath{{./figs/}}
\newcommand*\xbar[1]{%
    \hbox{%
		 \vbox{%
		 \hrule height 0.5pt % The actual bar
		 \kern0.5ex%         % Distance between bar and symbo
		 \hbox{%
		 \kern-0.1em%      % Shortening on the left side
		\ensuremath{#1}%
		\kern-0.1em%      % Shortening on the right side
		}%
		}%
		}%
		} 

\usepackage{algorithm} 
\usepackage{algpseudocode}

\algnewcommand{\Inputs}[1]{%
	\State \textbf{input:}
	\parbox[t]{.8\linewidth}{\raggedright #1}
}
\algnewcommand{\Initialize}[1]{%
	\State \textbf{initialization}
	\parbox[t]{.95\linewidth}{\raggedright #1}
}
\algnewcommand{\Outputs}[1]{%
	\State \textbf{output:}
	\parbox[t]{.8\linewidth}{\raggedright #1}
}
\begin{document}
\fontencoding{OT1}\fontsize{9.4}{11.25pt}\selectfont
\title{On In-network learning. A Comparative Study with Federated and Split Learning}
\author{Matei~Moldoveanu$^{\nmid}$ \qquad Abdellatif Zaidi$\:^{\dagger}$$\:^{\nmid}$\vspace{0.3cm}\\
$^{\dagger}$ Universit\'e Paris-Est, Champs-sur-Marne 77454, France\\
$^{\nmid}$ Mathematical and Algorithmic Sciences Lab., Paris Research Center, Huawei France \\
\{matei.catalin.moldoveanu@huawei.com, abdellatif.zaidi@u-pem.fr\}
}

\maketitle

\begin{abstract}
In this paper, we consider a problem in which distributively extracted features are used for performing inference in wireless networks. We elaborate on our proposed architecture, which we herein refer to as "in-network learning", provide a suitable loss function and discuss its optimization using neural networks. We compare its performance with both Federated- and Split learning; and show that this architecture offers both better accuracy and bandwidth savings. 
\end{abstract}
%\begin{IEEEkeywords}
%	Distributed learning, multiterminal statistical inference, edge learning.
%\end{IEEEkeywords}
\section{Introduction}
The unprecedented success of modern machine learning (ML) techniques in areas such as computer vision~\cite{obj_det_10_years_survey}, neuroscience~\cite{GLASER2019126}, image processing~\cite{medical_images_survey}, robotics~\cite{robotics_reinforced_learning} and natural language processing~\cite{VinyalsL15} has lead to an increasing interest for their application to wireless communication systems and networks over recent years. However, wireless networks have important intrinsic features which may require a deep rethinking of ML paradigms, rather than a mere adaptation of them. For example, while in traditional applications of ML the data used for training and/or inference is generally available at one point (centralized processing), it is typically highly distributed across the network in wireless communication systems. Examples include user localization based on received signals at base stations (BS)~\cite{CSI_fingerprinting,YLXKLTC20}, network anomaly detection and others. 

A prevalent approach would consist in collecting all data at one point (a cloud server) and then training a suitable ML model using all available data and processing power. This approach might not be appropriate in many cases, however, for it may require large bandwidth and network resources to share that data. In addition, applications such as autonomous vehicle driving might have stringent latency requirements that are incompatible with the principle of sharing data. In other cases, it might be desired not to share data for the sake of not infringing user privacy.
\begin{figure}[!hbpt]
	\centering
	\includegraphics[width=0.8\linewidth]{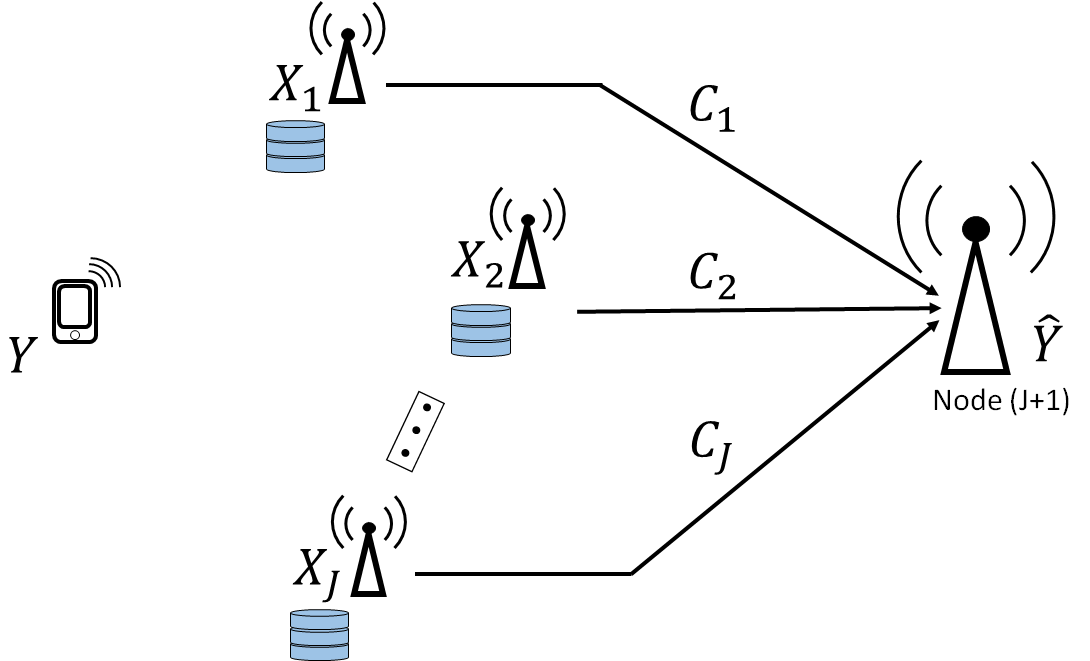}
	\caption{An example distributed inference problem.}
	\label{fig:prb_setting}
\end{figure}

A popular solution to the problem of learning distributively without sharing the data is the Federated learning (FL) of~\cite{mcmahan2017}. This architecture is most suitable for scenarios 
in which the training phase has to be performed distributively while the inference (or test) phase has to be performed centrally at one node. To this end, during the training phase nodes (e.g., BSs) that possess data are all equipped with copies of a single NN model which they simultaneously train on their available local data-sets. The learned weight parameters are then sent to a cloud- or parameter server (PS) which aggregates them, e.g. by simply computing their average. The process is repeated, every time re-initializing using the obtained aggregated model, until convergence. The rationale is that, this way, the model is progressively adjusted to account for all variations in the data, not only those of the local data-set. For recent advances on FL and applications in wireless settings, the reader may refer to~\cite{TBZNHC019,M.-AG20,YLQP20} and references therein. 

\begin{figure*}[!htpb]
	\begin{subfigure}[b]{0.5\linewidth}
		\centering
		\includegraphics[width=0.9\linewidth]{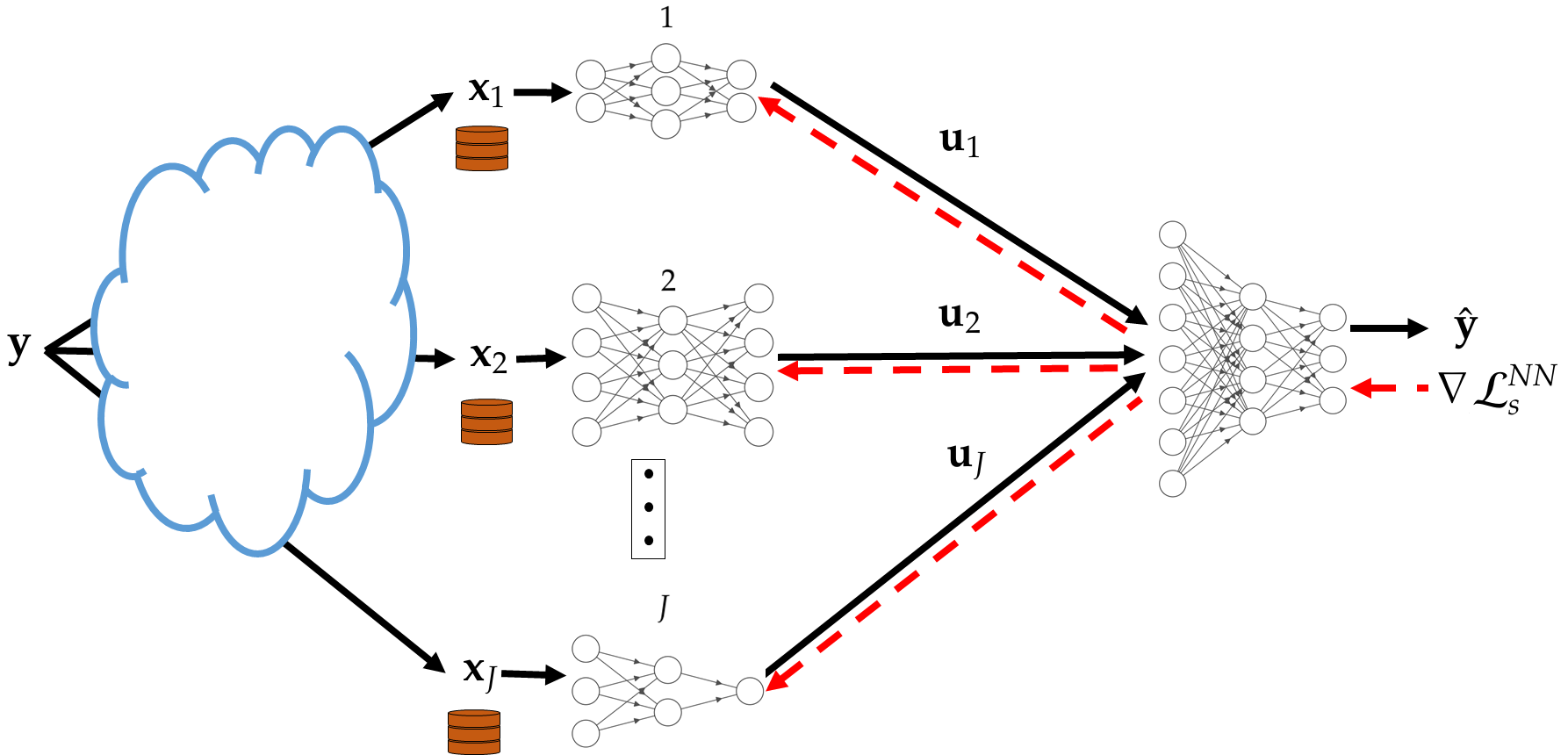}
		\caption{Training phase}
		\label{fig:prb_setting_train}
	\end{subfigure}
	\begin{subfigure}[b]{0.5\linewidth}
		\centering
		\includegraphics[width=0.9\linewidth]{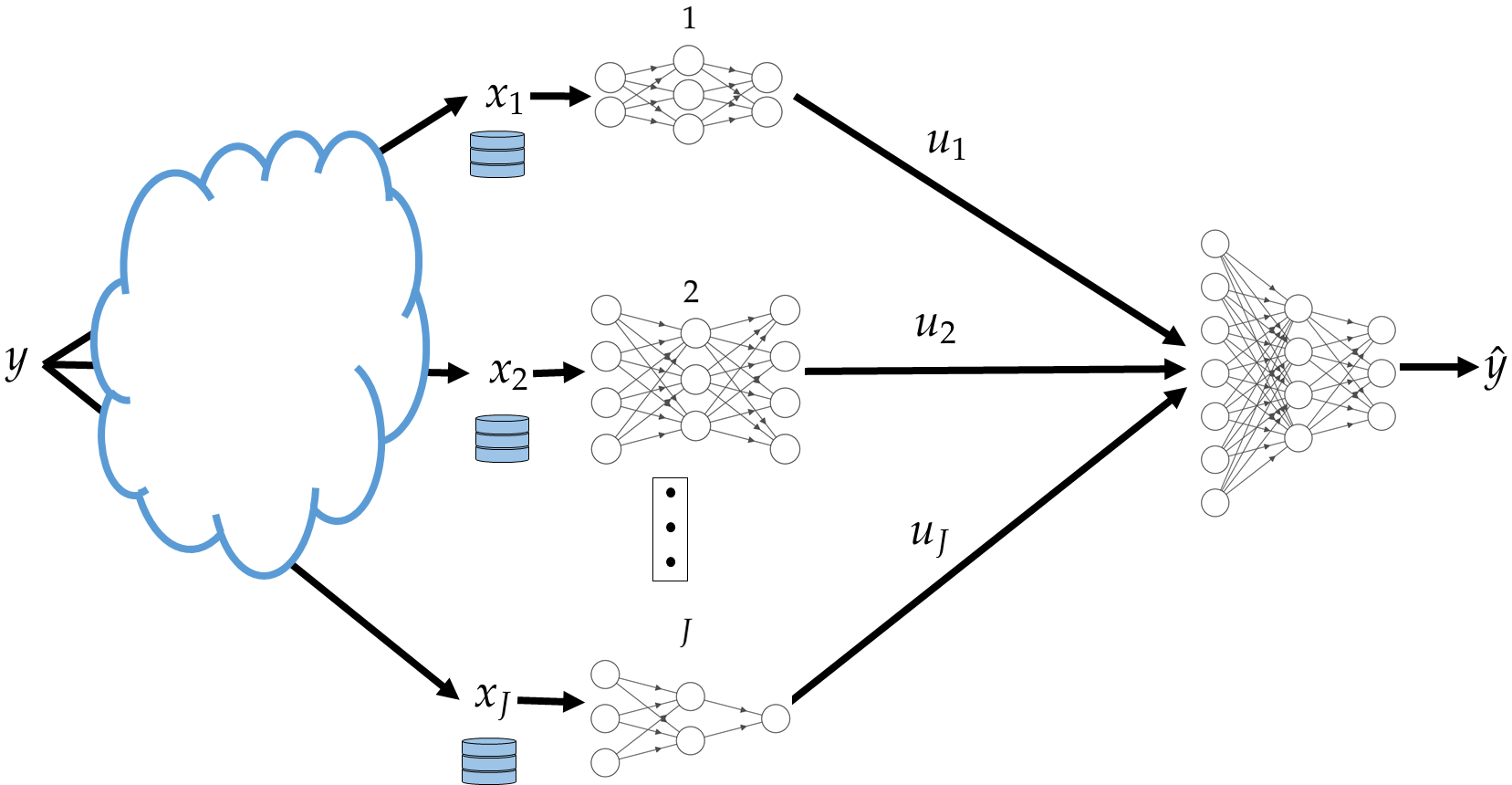}
		\caption{Inference phase}
		\label{fig:prb_setting_inf}
	\end{subfigure}
	\caption{In-network learning for the network model of Figure~\ref{fig:prb_setting}}
	\label{fig:multi_view_learning}
\end{figure*}

In this paper, we consider a different problem in which the processing needs to be performed distributively not only during the training phase as in FL but also during the inference or test phase. The model is  shown in Figure~\ref{fig:prb_setting}. In this problem, inference about a variable $Y$ (e.g., position of a user) needs to be performed at a distant central node (e.g., Macro BS), on the basis of summary information obtained from correlated measurements or signals $X_1,\hdots,X_J$ that are gotten at some proximity nodes (e.g., network edge BSs). Each of the edge nodes is connected with the central node via an error free link of given finite capacity. It is assumed that processing only (any) strict subset of the measurements or signals cannot yield the desired inference accuracy; and, as such, the $J$ measurements or signals $X_1,\hdots,X_J$ need to be processed during the inference or test phase (see Figure~\ref{fig:prb_setting_inf}).

The learning problem of Figure~\ref{fig:prb_setting} was first introduced and studied in~\cite{D-IB-discrete_gaussian} where a learning architecture which we name herein ``in-network (INL) learning", as well as a suitable loss function and a corresponding training algorithm, were proposed (see also~\cite{aguerri2019distributed, IB-problems}). The algorithm uses Markov sampling and is optimized using stochastic gradient descent. Also, multiple, possibly different, NN models are learned simultaneously, each at a distinct node.

In this paper, we study the specific setting in which edge nodes of a wireless network, that are connected to a central unit via error-free finite capacity links, implement the INL of~\cite{D-IB-discrete_gaussian,aguerri2019distributed}. We investigate in more details what the various nodes need to exchange during both the training and inference phase, as well as associated requirements in bandwidth. Finally, we provide a comparative study with (an adaptation of) FL and the Split Learning (SL) of~\cite{gupta2018distributed}.

\vspace{0.2cm}

\textbf{Notation:} Throughout, upper case letters  denote random variables, e.g., X;  lower case letters denote realizations of random variables, e.g., $x$; and calligraphic letters denote sets, e.g., $\mathcal{X}$. Boldface upper case letters denote vectors or matrices, e.g., $\dv X$. For random variables $(X_1,X_2,\hdots)$ and a set of integers $\mc J\subseteq \mathds{N}$, $X_{\mc J}$ denotes the set of variables with indices in  $\mc J$.

\section{Problem Formulation}

Consider the network inference problem shown in Figure~\ref{fig:prb_setting}. Here $J \geq 1$ nodes possess or can acquire data that is relevant for inference on a random variable $Y$. Let $\mc J =\{1, \hdots,J\}$ denote the set of such nodes. The inference on $Y$ needs to be done at some distant node (say, node $(J+1)$) which is connected to the nodes that possess raw data through error-free links of given finite capacities; and has to be performed without any sharing of raw data. The network may represent, for example, a wired network or a wireless mesh network operated in time or frequency division.

% For a given loss function $d(\cdot,\cdot)$ that measures discrepancies between true values of $Y$ and their estimated fits, what is the best precision for the estimation of $Y$ ? Clearly, discarding any of the relevant data $X_j$ can only lead to a reduced precision. Thus, intuitively features that collectively maximize information about $Y$ need to be extracted distributively by the nodes from the set $\mc J$, without explicit coordination between them; and they then need to propagate and combine appropriately at the node $(J+1)$. How should that be performed optimally without sharing raw data ? Furthermore, how should the information be fused optimally at node $(J+1)$ ?

More formally, the processing at node $j \in \mc J$ is a mapping
\begin{equation}
\phi_j: \mc X_j \longrightarrow [1:2^{C_j}];
\label{encoding-function-sender-node-j}
\end{equation} 
and that at node $(J+1)$ is a mapping
\begin{equation}
\psi: [1:2^{C_1}] \times \hdots \times [1:2^{C_J}] \longrightarrow \hat{\mc Y}.
\label{decoding-function-final-decision-node}
\end{equation} 
In this paper, we choose the reconstruction set $\hat{\mc Y}$ to be the set of distributions on $\mc Y$, i.e., $\hat{\mc Y} = \mc P(\mc Y)$; and we measure discrepancies between true values of $Y \in \mc Y$ and their estimated fits in terms of average logarithmic loss, i.e., for $(y, \hat{P}) \in \mc Y \times \mc P(\mc Y)$
\begin{equation}
d(y, \hat{P}) = \log \frac{1}{\hat{P}(y)}.
\label{definition-log-loss-distorsion-measure}
\end{equation}
As such the performance of a distributed inference scheme $\left( (\phi_j)_{j \in \mc J}, \psi \right)$ is evaluated as
\begin{equation}
\Delta = H(Y) - \mathbb{E}\left[d(Y,\hat{Y})\right].
\label{relevance-measure-at-end-decision-node}
\end{equation}
\noindent In practice, in a supervised setting, the mappings given by~\eqref{encoding-function-sender-node-j} and~\eqref{decoding-function-final-decision-node} need to be learned from a set of training data samples $\{(x_{1,i},\hdots,x_{J,i},y_i)\}_{i=1}^n$. The data is distributed such that the samples $\dv x_j :=(x_{j,1}\hdots,x_{j,n})$ are available at node $j$ for $j \in \mc J$ and the desired predictions $\dv y :=(y_1 \hdots y_n)$ are available at node $(J+1)$.

\vspace{-0.3cm}
\section{In-network Learning}\label{propsol}

\vspace{-0.1cm}

We parametrize the possibly stochastic mappings~\eqref{encoding-function-sender-node-j} and~\eqref{decoding-function-final-decision-node} using neural networks. This is depicted in Figure~\ref{fig:multi_view_learning}. The NNs at the various nodes are arbitrary and can be chosen independently -- for instance, they need not be identical as in FL. It is only required that the following mild condition, which as will become clearer from what follows facilitates the back-propagation, be met 
\begin{equation}
\sum_{j=1}^J (\text{Size of last layer of NN $j$}) = \text{Size of first layer of NN (J+1)}.
\label{condition-concatenation-activations-vectors}
\end{equation}
A possible suitable loss function was shown to be given by~\cite{aguerri2019distributed}
\begin{align}
	& \mc L^{\text{NN}}_s(n) =  \frac{1}{n} \sum_{i=1}^n \log Q_{\boldsymbol{\phi_{\mc J}}}(y_i|u_{1,i},\hdots,u_{J,i}) \nonumber\\
& \qquad + \frac{s}{n} \sum_{i=1}^n \sum_{j=1}^J \left( \log Q_{\boldsymbol{\phi_j}}(y_i|u_{j,i}) - \log \left(\frac{P_{\boldsymbol{\theta_j}}(u_{j,i}|x_{j,i})}{Q_{\boldsymbol{\varphi_j}}(u_{j,i})}\right) \right),
\label{loss function}
\end{align}
where $s$ is a Lagrange parameter and for $j \in \mc J$ the distributions $P_{\boldsymbol{\theta_j}}(u_j|x_j)$, $Q_{\boldsymbol{\phi_j}} (y|u_j)$, $Q_{\boldsymbol{\phi_{\mc J}}}(y|u_{\mc J})$ are variational ones whose parameters are determined by the chosen NNs using the re-parametrization trick of~\cite{kingma2013auto}; and $Q_{\boldsymbol{\varphi_j}}(u_{j})$ are priors known to the encoders. For example, denoting by $f_{\theta_j}$ the NN used at node $j \in \mc J$ whose (weight and bias) parameters are given by $\boldsymbol{\theta}_j$, for regression problems the conditional distribution $P_{\boldsymbol{\theta_j}}(u_j|x_j)$ can be chosen to be multivariate Gaussian, i.e., $P_{\boldsymbol{\theta_j}}(u_j|x_j) = \mc {N} (u_j; \boldsymbol{\mu}_j^{\theta}, \dv \Sigma_j^{\theta})$. For discrete data, concrete variables (i.e., Gumbel-Softmax) can be used instead. 

% {Similarly, for decoders $Q_{Y|U_k}$ over $\mc Y$ for each element on $\mc U_k$, and the decoding distribution $Q_{Y|U_{\mc K}}$ over $\mc Y$ for each element in $ \mc U_1\times \cdots \times \mc U_K$, let $Q_{\psi_k}(y|u_k)$, $Q_{\psi_{\mc K}}(y|u_{\mc K})$ denote the families of distributions parametrized by the output of the DNNs $f_{\psi_k}, f_{\psi_{\mc K}}$, respectively. Finally, for the prior distributions $Q_{U_k}(u_k)$ over $\mc U_k$ we define the family of distributions $Q_{\varphi_k}(u_k)$, which do not depend on a DNN.}

The rationale behind the choice of loss function~\eqref{loss function} is that in the regime of large $n$, if the encoders and decoder are not restricted to use NNs under some conditions~\footnote{The optimality is proved therein under the assumption that for every subset $\mc S \subseteq \mc J$ it holds that $X_{\mc S} \mkv Y \mkv X_{{\mc S}^c}$. The RHS of~\eqref{optimal loss function} is achievable for arbitrary distributions, however, regardless of such an assumption.} the optimal stochastic mappings $P_{U_j|X_j}$, $P_U$, $P_{Y|U_j}$ and $P_{Y|U_{\mc J}}$ are found by marginalizing the joint distribution that maximizes the following Lagrange cost function~\cite[Proposition 2]{aguerri2019distributed}
\begin{equation}
\mathcal L^{\mathrm{optimal}}_s = - H(Y|U_{\mc J}) - s \sum_{j=1}^J \Big[H(Y|U_j)+I(U_j;X_j)\Big].
\label{optimal loss function}
\end{equation}
where the maximization is over all joint distributions of the form $P_{Y}\prod_{j=1}^JP_{X_j|Y}\prod_{j=1}^J P_{U_j|X_j}$.

\subsection{Training Phase}

During the forward pass, every node $j \in \mc J$ processes mini-batches of size, say, $b_j$ of its training data-set $\dv x_j$. Node $j \in \mc J$ then sends a vector whose elements are the activation values of the last layer of (NN $j$). Due to~\eqref{condition-concatenation-activations-vectors} the activation vectors are concatenated vertically at the input layer of NN (J+1). The forward pass continues on the NN (J+1) until the last layer of the latter.

The parameters of NN (J+1) are updated using standard backpropgation. Specifically, let $L_{J+1}$ denote the index of the last layer of NN $(J+1)$. Also, let, for $l \in [2:L_{J+1}]$, 
$\bm{w}_{J+1}^{[l]}$, $\bm{b}_{J+1}^{[l]}$ and $\bm{a}_{J+1}^{[l]}$ denote respectively the weights, biases and activation values at layer $l$ for the NN $(J+1)$; and $\sigma$ is the activation function. Node $(J+1)$ computes the error vectors
\begin{subequations}
\begin{align}
\label{back_prop_out_layer}
& \boldsymbol{\delta}_{J+1}^{[L_{J+1}]} =\nabla_{\bm a_{J+1}^{[L_{J+1}]}}\mc L^{NN}_s(b) \odot \sigma'(\bm{w}_{J+1}^{[L_{J+1}]}\bm{a}_{f}^{[L_{(J+1)}-1]}+\bm{b}_{J+1}^{[L_{J+1}]}) \\
& \boldsymbol{\delta}_{J+1}^{[l]} =[(\bm{w}_{J+1}^{[l+1]})^{T} \boldsymbol{\delta}_{J+1}^{[l+1]}]\odot \sigma'(\bm{w}_{J+1}^{[l]}\bm{a}_{J+1}^{[l-1]}+\bm{b}_{J+1}^{[l]})\:\: \forall \:\: l \in [2,L_{J+1}-1]% \forall was initially \text{for}
\label{back_prop_err}\\
& \boldsymbol{\delta}_{J+1}^{[1]} =[(\bm{w}_{J+1}^{[2]})^{T} \boldsymbol{\delta}_{J+1}^{[2]}],\label{back_prop_err_first_layer}
\end{align} 
\label{equations-backpropagation}
\end{subequations}
and then updates its weight- and bias parameters as
\begin{subequations}
\begin{align}
\bm{w}^{[l]}_{J+1} &\rightarrow \bm{w}_{J+1}^{[l]}-\eta \boldsymbol{\delta}_{J+1}^{[l]}(\bm{a}_{J+1}^{[l-1]})^T,\label{back_prop_bias}\\
\bm{b}^{[l]}_{J+1}&\rightarrow \bm{b}_{J+1}^{[l]}-\eta \boldsymbol{\delta}_{J+1}^{[l]},\label{back_prop_weight}
\end{align}
\label{equations-parameters-update}
\end{subequations}
where $\eta$ designates the learning parameter~\footnote{For simplicity $\eta$ and $\sigma$ are assumed here to be identical for all NNs.}.

\begin{figure}[!hbpt]
	\centering
	\includegraphics[width=1\linewidth]{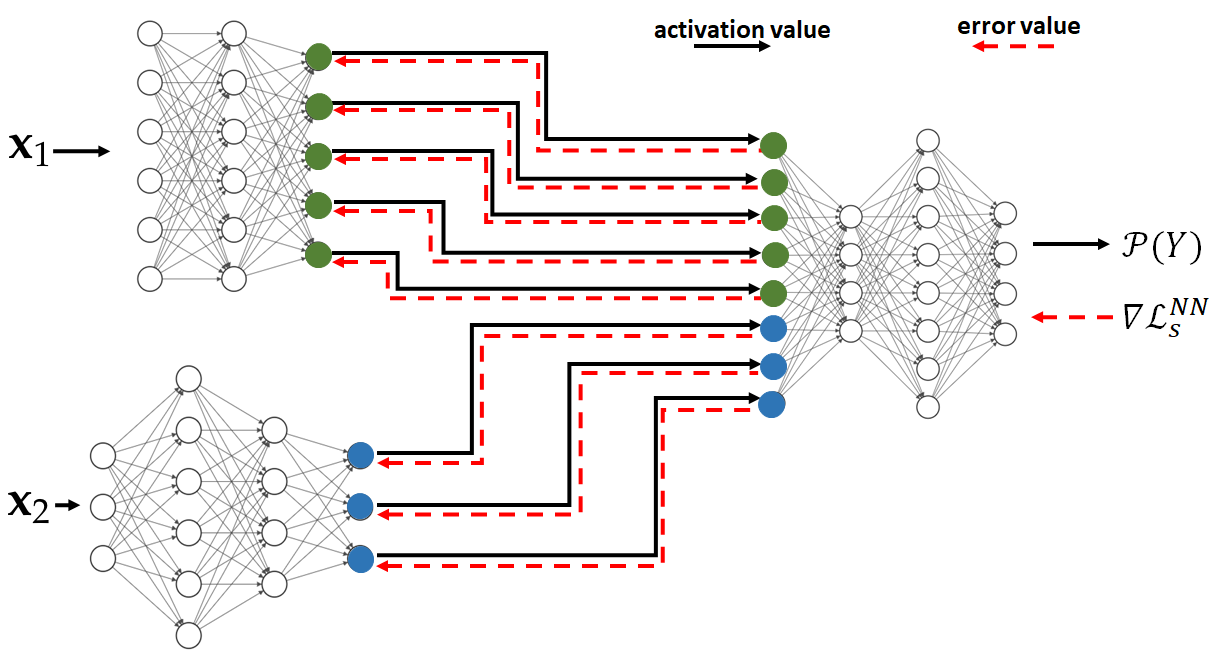}
	\caption{Illustration of the Forward and Backward passes for an example in-network learning with $J=2$.}
	\label{fig_2_node_mapping}
\end{figure}

\begin{remark}
It is important to note that for the computation of the RHS of~\eqref{back_prop_out_layer} node $(J+1)$, which knows $Q_{\phi_{\mc J}}(y_i|u_{1i},\hdots,u_{Ji})$ and $Q_{\phi_j}(y_i|u_{ji})$ for all $i \in [1:n]$ and all $j \in \mc J$, only the derivative of $\mc L^{\text{NN}}_{s}(n)$ w.r.t. the avtivation vector $\dv a^{L_{J+1}}_{J+1}$ is required. For instance, node $(J+1)$ does not need to know any of the conditional variationals $P_{\boldsymbol{\theta_j}}(u_j|x_j)$ or the priors $Q_{\boldsymbol{\varphi_j}}(u_j)$.
\end{remark}

The backward propagation of the error vector from node $(J+1)$ to the nodes $j$, $j=1,\hdots,J$, is as follows. Node $(J+1)$ splits horizontally the error vector of its input layer into $J$ sub-vectors with sub-error vector $j$ having size $L_j$, the dimension of the last layer of NN $j$ [recall~\eqref{condition-concatenation-activations-vectors} and that the activation vectors are concatenated vertically during the forward pass]. See Figure~\ref{fig_2_node_mapping}. The backward propagation then continues on each of the $J$ input NNs simultaneously, each of them essentially applying operations similar to~\eqref{equations-backpropagation} and~\eqref{equations-parameters-update}. 

\begin{remark}
Let $\boldsymbol{\delta}_{J+1}^{[1]}(j)$ denote the sub-error vector sent back from node $(J+1)$ to node $j \in \mc J$. It is easy to see that, for every $j \in \mc J$, 
	\begin{equation}
		\nabla_{\bm a^{L_j}_{j}}\mc L^{NN}_s(b_j)=\boldsymbol{\delta}_{J+1}^{[1]}(j)-s\nabla_{\bm a^{L_j}_{j}}\left(\sum_{i=1}^b \log \left(\frac{P_{\boldsymbol{\theta_j}}(u_{j,i}|x_{j,i})}{Q_{\boldsymbol{\varphi_j}}(u_{j,i})}\right)\right);
	\end{equation}
and this explains why node $j \in \mc J$ needs only the part $\boldsymbol{\delta}_{J+1}^{[1]}(j)$, not the entire error vector at node $(J+1)$.
\end{remark}

\subsection{Inference Phase}

During this phase node $j$ observes a new sample $x_j$. It uses its NN to output an encoded value $u_j$ which it sends to the decoder. After collecting $(u_1, \cdots, u_J)$ from all input NNs, node $(J+1)$ uses its NN to output an estimate of $Y$ in the form of soft output $Q_{\phi_{\mc J}}(Y|u_1,\hdots, u_J)$. The procedure is depicted in Figure~\ref{fig:prb_setting_inf}.

\begin{remark}
A suitable practical implementation in wireless settings can be obtained using Orthogonal Frequency Division Multiplexing (OFDM). That is, the $J$ input nodes are allocated non-overlapping bandwidth segments and the output layers of the corresponding NNs are chosen accordingly. The encoding of the activation values can be done, e.g., using entropy type coding~\cite{FHH-L21}. 
\end{remark}

\subsection{Bandwidth requirements}\label{bandreq}

In this section, we study the requirements in bandwidth of our in-network learning. Let $q$ denote the size of the entire data set (each input node has a local dataset of size $\frac{q}{J}$),  $p=L_{J+1}$ the size of the input layer of NN $(J+1)$ and $s$ the size in bits of a parameter. Since as per~\eqref{condition-concatenation-activations-vectors}, the output of the last layers of the input NNs are concatenated at the input of NN $(J+1)$ whose size is $p$, and each activation value is $s$ bits, one then needs $\dfrac{2sp}{J}$ bits for each data point -- the factor $2$ accounts for both the forward and backward passes; and, so, for an epoch our in-network learning requires $\dfrac{2pqs}{J}$ bits. 

Note that the bandwidth requirement of in-network learning does not depend on the sizes of the NNs used at the various nodes, but does depend on the size of the dataset. For comparison, notice that  with FL one would require $2NJs$, where $N$ designates the number of (weight- and bias) parameters of a NN at one node. For the SL of~\cite{gupta2018distributed}, assuming for simplicity that the NNs $j=1,\hdots,J$ all have the same size $\eta N$, where $\eta \in [0,1]$, SL requires $(2pq+\eta NJ)s$ bits for an entire epoch.

 The bandwidth requirements of the three schemes are summarized and compared in Table~\ref{band_res} for two popular neural networks, VGG16 ($N=138,344,128$ parameters) and ResNet50 ($N=25,636,712$ parameters) and two example datsets, $q =50, 000$ data points and $q=500, 000$ data points. The numerical values are set as $J=500$, $p=25088$ and $\eta=0.88$ for ResNet50 and $0.11$ for VGG16. 

\begin{table}[hb]
	\centering
	
	\begin{tabular}{|c|l|l|l|}
		\hline
		& \thead{Federated\\learning} & \thead{Split\\learning} & \thead{In-network\\learning} \\ \hline
		Bandwidth requirement  & $	2NJs$          &    $\left(2pq+\eta NJ\right)s$            &   \thead{$\dfrac{2pqs}{J}$}    \\ \hline
	    \thead{VGG 16  \\ 50,000 data points}   &   4427 Gbits  &  324 Gbits     & 0.16 Gbits     \\ \hline
		\thead{ResNet 50  \\ 50,000 data points}	&   820 Gbits    &  441 Gbits & 0.16 Gbits     \\ \hline
		\thead{VGG 16  \\ 500,000 data points}  		&   4427 Gbits      &   1046 Gbits   & 1.6 Gbits           \\ \hline
		\thead{ResNet 50 \\ 500,000 data points} 		&   820 Gbits   &   1164 Gbits   & 1.6 Gbits            \\ \hline
	\end{tabular}
\caption{Bandwidth requirements of INL, FL and SL.}
\label{band_res}
\end{table}

\vspace{-0.4cm}
\section{Experimental Results}

We perform two series of experiments. In both cases, the used dataset is the CIFAR-10 and there are five client nodes.

\subsection{Experiment 1}
In this setup, we create five sets of noisy versions of the images of CIFAR-10. To this end, the CIFAR images are first normalized, and then corrupted by additive Gaussian noise with standard deviation set respectively to $0.4, 1, 2, 3, 4$. 

%\begin{figure}[htpb]
%	\centering
%	\includegraphics[width=1\linewidth]{noisy_pictures_example.png}
%	\caption{Example of an image with different levels of added noise}
%	\label{fig:image_corr_noise}
%\end{figure}

\begin{figure}[htpb]
	\centering
	\includegraphics[width=1\linewidth]{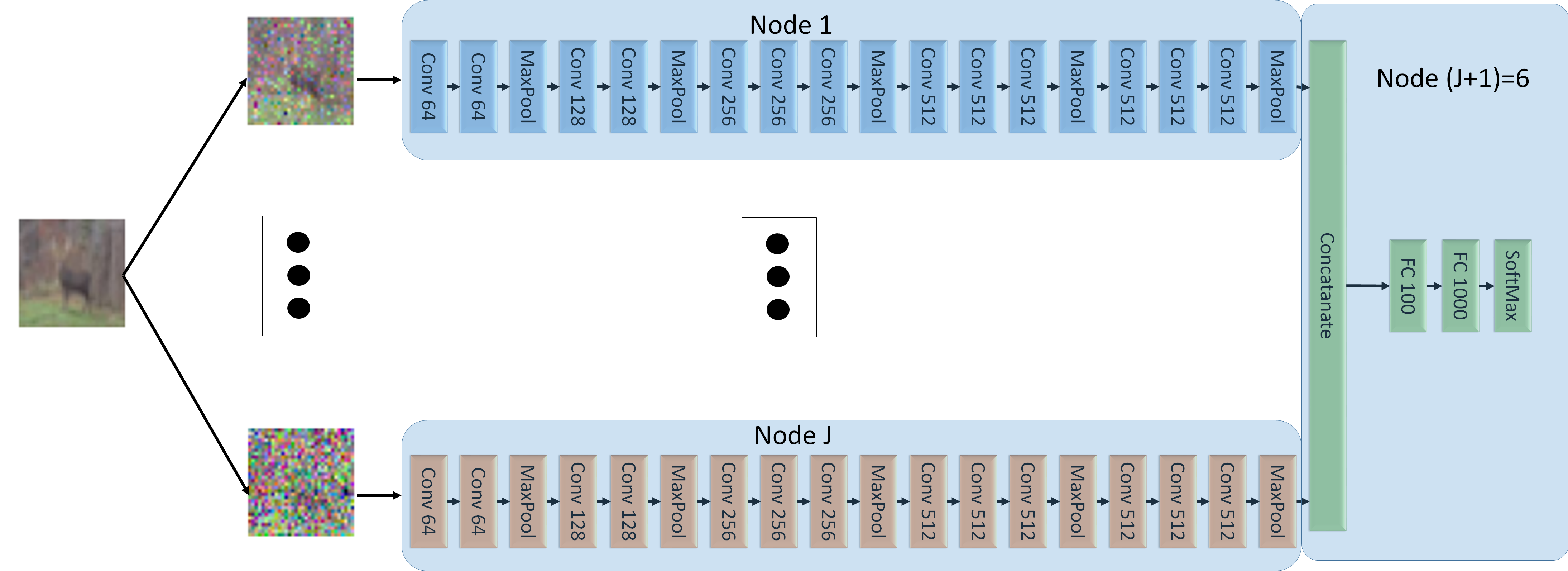}
	\caption{Network architecture. \textit{Conv} stands for a convolutional layer, \textit{Fc} stand for a fully connected layer.}
	\label{fig:nt_arh_5}
\end{figure}

For our INL each of the five input NNs is trained on a different noisy version of the same image. Each NN uses a variation of the VGG network of~\cite{vgg_small_data}, with the categorical cross-entropy as the loss function, L2 regularization, and Dropout and BatchNormalization layers. Node $(J+1)$ uses two dense layers. The architecture is shown in Figure~\ref{fig:nt_arh_5}. In the experiments, all five (noisy) versions of every CIFAR-10 image are processed simultaneously, each by a different NN at a distinct node, through a series of convolutional layers. The outputs are then concatenated and then passed through a series of dense layers at node $(J+1)$. 

For FL, each of the five client nodes is equipped with the \textit{entire} network of Figure~\ref{fig:nt_arh_5}. The dataset is split into five sets of equal sizes; and the split is now performed such that all five noisy versions of a same CIFAR-10 image are presented to the same client NN (distinct clients observe different images, however). For SL of~\cite{gupta2018distributed}, each input node is equipped with an NN formed by \textit{all} fives branches with convolution networks (i.e., all the network of Fig.~\ref{fig:nt_arh_5}, except the part at Node $(J+1)$);  and node $(J+1)$ is equipped with fully connected layers at Node $(J+1)$ in Figure~\ref{fig:nt_arh_5}. Here, the processing during training is such that each input NN concatenates vertically the outputs of all convolution layers and then passes that to node $(J+1)$, which then propagates back the error vector. After one epoch at one NN, the learned weights are passed to the next client, which performs the same operations on its part of the dataset.

\vspace{0.2cm}

\begin{figure}[!htpb]
\begin{subfigure}{0.9\linewidth}
\centering
	\includegraphics[width=1\linewidth]{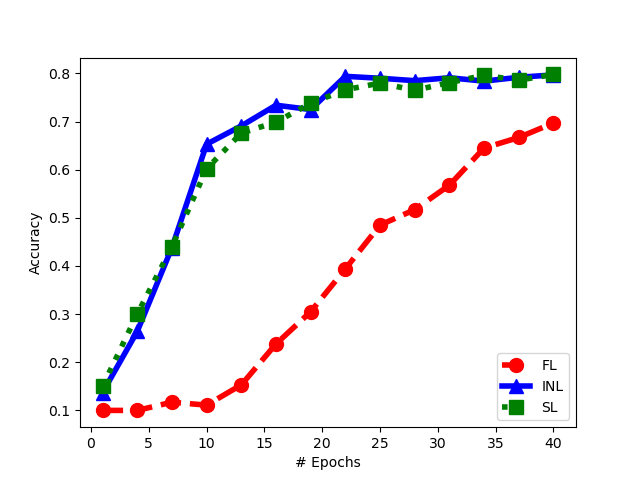}
	\caption{Accuracy vs. $\#$ of epochs.}
	\label{fig:accvsepochexp1}
	\end{subfigure}
	\hfill
	\begin{subfigure}{0.9\linewidth}
	\centering
		\includegraphics[width=\linewidth]{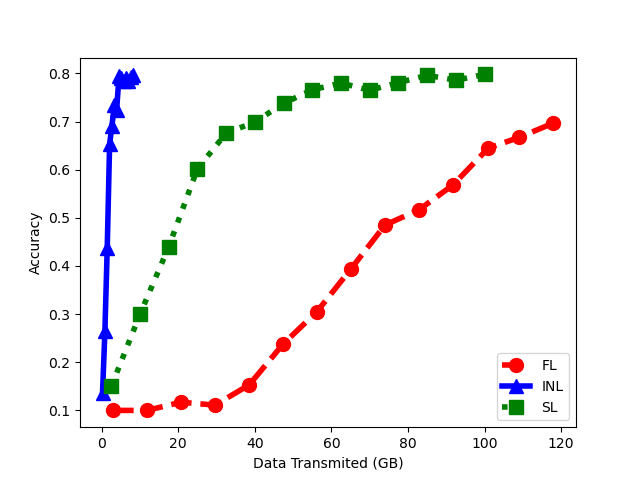}
	\caption{Accuracy vs. bandwidth cost.}
	\label{fig:accvsdataexp1}
	\end{subfigure}
		\caption{Comparison of INL, FL and SL for Experiment 1}
	\label{fig:experiment_results_1}
\end{figure}

Figure~\ref{fig:accvsepochexp1} depicts the evolution of the classification accuracy on CIFAR-10 as a function of the number of training epochs, for the three schemes. As visible from the figure, the convergence of FL is relatively slower comparatively. Also the final result is less accurate. Figure~\ref{fig:accvsdataexp1} shows the amount of data needed to be exchanged among the nodes (i.e., bandwidth resources) in order to get a prescribed value of classification accuracy. Observe that both our INL and SL require significantly less data exchange than FL; and our INL is better than SL especially for small values of bandwidth.

\vspace{-0.2cm}

\begin{figure}[!h]
	\centering
	\includegraphics[width=1\linewidth]{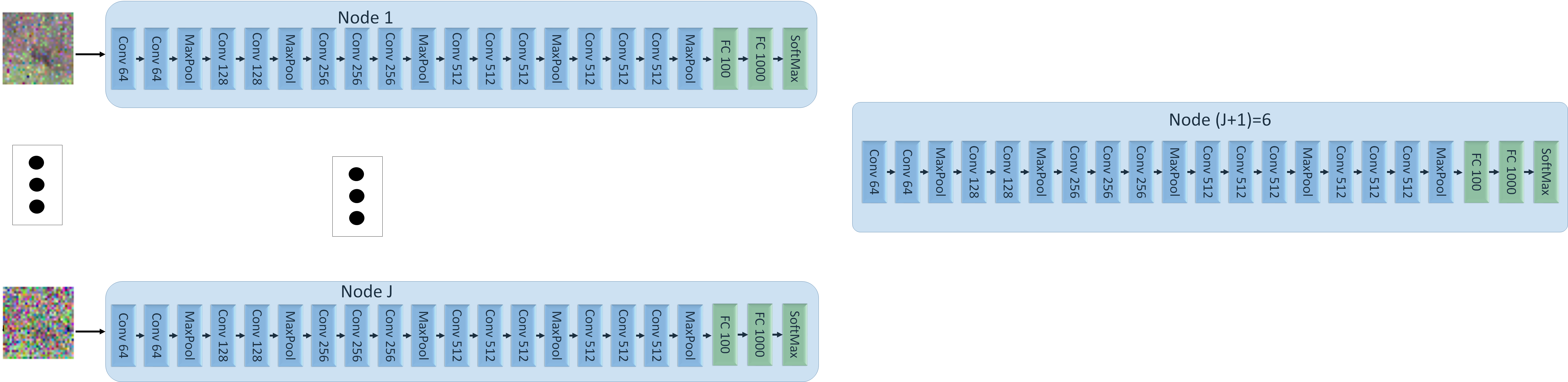}
	\caption{Used NN architecture for FL in Experiment 2} 
	\label{fig:accvsdata}
\end{figure}

\iffalse
\begin{figure}[htpb]
	\centering
	\includegraphics[width=1\linewidth]{results_acc_simple_pretty_v5-V2.png}
	\caption{Accuracy versus number of epochs.}
	\label{fig:accvsepoch}
\end{figure}

\begin{figure}[htpb]
	\centering
	\includegraphics[width=1\linewidth]{accvsdataV2.png}
	\caption{Accuracy versus bandwidth cost.}
	\label{fig:accvsdata}
\end{figure}
\fi

\vspace{-0.5cm}

\subsection{Experiment 2}

In Experiment 1, the entire training dataset was partitioned differently for INL, FL and SL (in order to account for the particularities of the three). In this second experiment, they are all trained on the same data. Specifically, each client NN sees all CIFAR-10 images during training; and its local dataset differs from those seen by other NNs only by the amount of added Gaussian noise (standard deviation chosen respectively as $0.4, 1, 2, 3, 4$). Also, for the sake of a fair comparison between INL, FL and SL the nodes are set to utilize fairly the same NNs for the three of them (see, Fig.~\ref{fig:accvsdata}).
 
% Here the setup is similar to the above, except that for FL each of the five client nodes is now  equipped with an NN formed of one branch with convolution networks (i.e. node 1 of Fig.~\ref{fig:nt_arh_5}) connected to the fully connected layers of node $(J+1)$. For SL each of the five client nodes has the same NN as the INL case, while node (J+1) has an NN formed of only the fully connected layers from Fig.~\ref{fig:nt_arh_5}, without the concatenate layer. For all three cases each input node has access to a different noisy version of the same image. The evaluation of the NN trained by SL and FL is done on a dataset obtained by averaging the five noisy versions of each image.

\vspace{-0.4cm}

\begin{figure}[!htpb]
	\begin{subfigure}{0.9\linewidth}
		\centering
		\includegraphics[width=1\linewidth]{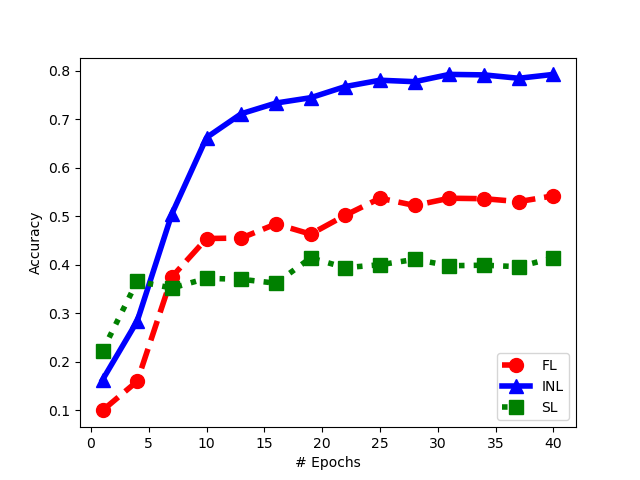}
		\caption{Accuracy vs. $\#$ of epochs.}
		\label{fig:accvsepochexp2}
	\end{subfigure}
	\hfill
	\begin{subfigure}{0.9\linewidth}
		\centering
		\includegraphics[width=1\linewidth]{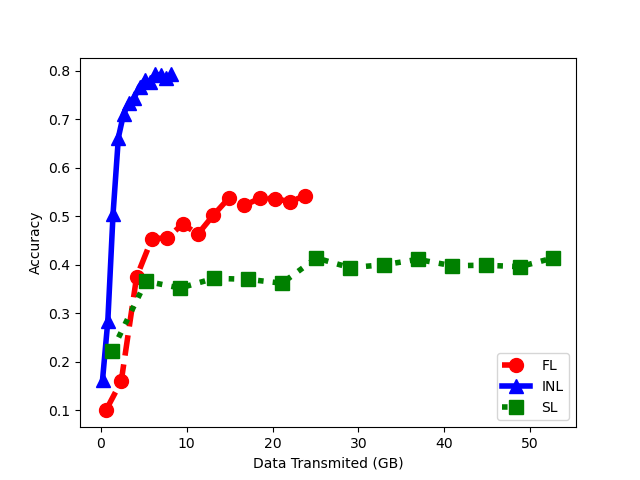}
		\caption{Accuracy vs. bandwidth cost.}
		\label{fig:accvsdataexp2}
	\end{subfigure}
	\caption{Comparison of INL, FL and SL for Experiment 2.}
	\label{fig:experiment_results_2}
\end{figure}

\vspace{0.0cm}

Figure~\ref{fig:accvsdataexp2} shows the performance of the three schemes during the inference phase in this case (for FL the inference is performed on an image which has average quality of the five noisy input images for INL and SL). Again, observe the benefits of INL over FL and SL in terms of both achieved accuracy and bandwidth requirements. 

\begin{remark}
INL has desirable features, among which that it is easily amenable to extensions to arbitrary networks, including networks that involve hops. This will be reported elsewhere.
\end{remark}

\iffalse
\begin{figure}[htpb]
	\centering
	\includegraphics[width=1\linewidth]{FL_exp_1.png}
	\caption{FL setup for Experiment 1.}
	\label{fig:accvsdata}
\end{figure}
\begin{figure}[htpb]
	\centering
	\includegraphics[width=1\linewidth]{SL_exp_2.png}
	\caption{SL setup for experiment 2.}
	\label{fig:accvsdata}
\end{figure}
\fi

\iffalse
\begin{figure}[htpb]
	\centering
	\includegraphics[width=1\linewidth]{SL_exp_1.png}
	\caption{SL setup for Experiment 1.}
	\label{fig:accvsdata}
\end{figure}
\fi

\vspace{-0.4cm}

%================================================================================================
%\newpage
%\appendix
%\input{appendices-draft-spawc.tex}

%-------------------------------------------------------------------------------------
\bibliographystyle{IEEEtran}
\bibliography{IEEEabrv,references} 

\end{document}